\pdfoutput=1

\documentclass[11pt]{article}

\usepackage[final]{acl}

\usepackage{times}
\usepackage{latexsym}

\usepackage[T1]{fontenc}

\usepackage[utf8]{inputenc}

\usepackage{microtype}

\usepackage{inconsolata}

\usepackage{graphicx}

\title{Efficient ASR for Low-Resource Languages: Leveraging Cross-Lingual Unlabeled Data}

\author{
  \textbf{Srihari Bandarupalli\textsuperscript{1*}},
  \textbf{Bhavana Akkiraju\textsuperscript{1*}},
  \textbf{Charan Devarakonda\textsuperscript{1}},\\
  \textbf{Vamsiraghusimha Narsinga\textsuperscript{1}},
  \textbf{Anil Kumar Vuppala\textsuperscript{1}}\\
  \textsuperscript{1}Speech Processing Lab, International Institute of Information Technology Hyderabad, India\\
  \texttt{\{srihari.bandarupalli,bhavana.akkiraju,sricharan.d\}@research.iiit.ac.in}\\
  \texttt{narasinga.vamshi@research.iiit.ac.in, anil.vuppala@iiit.ac.in}\\
  \small{* These authors contributed equally.}
}

\begin{document}
\maketitle
\begin{abstract}
Automatic speech recognition for low-resource languages remains fundamentally constrained by the scarcity of labeled data and computational resources required by state-of-the-art models. We present a systematic investigation into cross-lingual continuous pretraining for low-resource languages, using Perso-Arabic languages (Persian, Arabic, and Urdu) as our primary case study. Our approach demonstrates that strategic utilization of unlabeled speech data can effectively bridge the resource gap without sacrificing recognition accuracy. We construct a 3,000-hour multilingual corpus through a scalable unlabeled data collection pipeline and employ targeted continual pretraining combined with morphologically-aware tokenization to develop a 300M parameter model that achieves performance comparable to systems 5 times larger. Our model outperforms Whisper Large v3 (1.5B parameters) on Persian and achieves competitive results on Arabic and Urdu despite using significantly fewer parameters and substantially less labeled data. These findings challenge the prevailing assumption that ASR quality scales primarily with model size, revealing instead that data relevance and strategic pretraining are more critical factors for low-resource scenarios. This work provides a practical pathway toward inclusive speech technology, enabling effective ASR for underrepresented languages without dependence on massive computational infrastructure or proprietary datasets. To encourage further research, we release our entire codebase and model checkpoints\footnote{\url{https://github.com/sriharib128/EfficientASR.git}}.
\end{abstract}

\section{Introduction}

Accurate ASR for morphologically complex, low-resource languages remains a pressing challenge. Recent self-supervised methods like Wav2Vec 2.0~\cite{baevski2020wav2vec20frameworkselfsupervised} and HuBERT~\cite{hsu2021hubertselfsupervisedspeechrepresentation} have advanced ASR using large-scale unlabeled data. Building on this, multilingual models like XLS-R~\cite{babu2021xlsrselfsupervisedcrosslingualspeech} have improved cross-lingual ASR with diverse pretraining~\cite{wang-etal-2021-voxpopuli, Pratap2020MLSAL, ardila-etal-2020-common, valk2020voxlingua107datasetspokenlanguage, babel}. However, these gains mainly benefit high-resource languages, while low-resource ones still suffer from limited labeled data.

For languages like Persian, Arabic, and Urdu, the intersection of limited labeled data, complex orthographies, and rich morphology poses considerable difficulties for ASR. Recent systematic evaluations of state-of-the-art models for Persian\footnote{\url{https://huggingface.co/spaces/navidved/open_persian_asr_leaderboard}}, Arabic~\cite{wang2024openuniversalarabicasr}, and Urdu~\cite{urdu_leader} indicate that models like Whisper Large v3~\cite{whisper} and Seamless Large v2~\cite{communication2023seamlessmultilingualexpressivestreaming} exhibit superior performance. However, they are computationally intensive (1.5B+ parameters), making efficient deployment challenging. Some prior works~\cite{getman24_interspeech} have explored continuous pretraining, but their analysis focuses on high-resource scenarios and single-language improvements.


In this work, we address these gaps by systematically investigating ASR for low-resource languages, using Perso-Arabic languages as our primary case study. Our approach combines scalable unlabeled data collection with targeted continuous pretraining and morphologically-aware tokenization to demonstrate that compact models (300M parameters) can achieve performance comparable to systems five times larger. Through systematic evaluation across Persian, Arabic, and Urdu, we provide empirical evidence that strategic utilization of cross-lingual unlabeled data can effectively overcome resource constraints without sacrificing recognition accuracy.

\begin{figure*}[t]
  \centering
  \includegraphics[width=\linewidth]{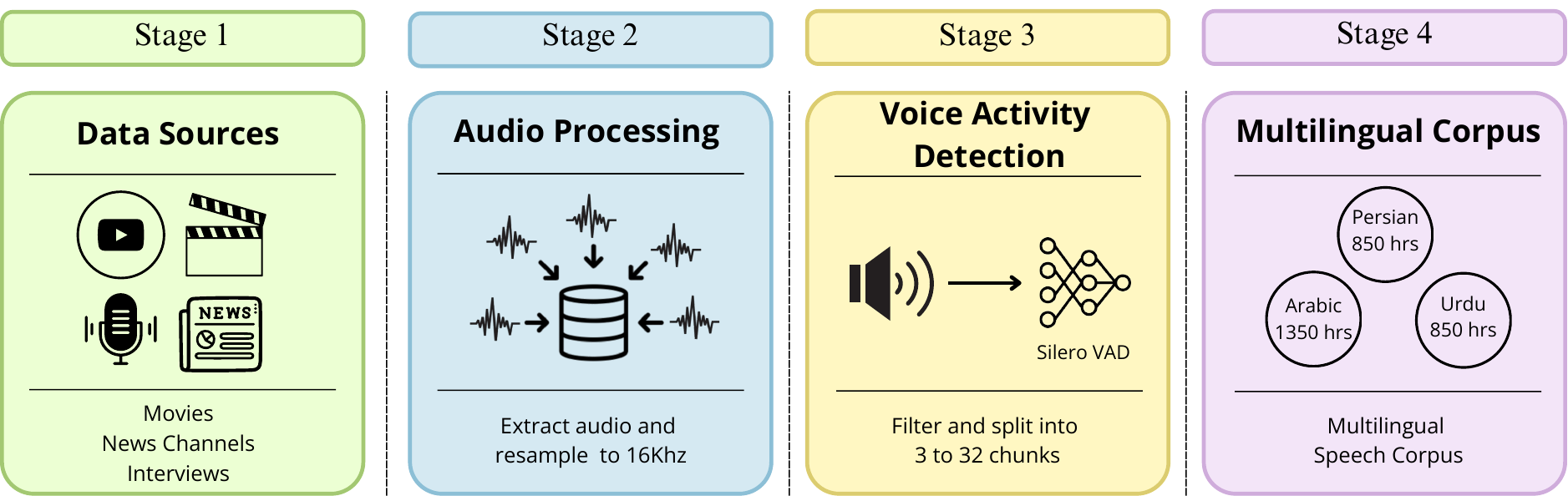}
  \caption{Systematic pipeline for constructing a robust, multilingual unlabeled speech corpus.}
  \vspace{-7.5pt}
  \label{fig:Unlabelled_Pipeline}
\end{figure*}

\vspace{-3 pt}

\section{Data}
\label{sec:Data}

We systematically constructed both unlabeled and labeled speech corpora using a modular data infrastructure to enable controlled investigation of cross-lingual ASR for morphologically-complex, low-resource languages.

\subsection{Unlabeled Corpus for Robust Pretraining}

We developed a scalable pipeline (Figure~\ref{fig:Unlabelled_Pipeline}) to curate high-quality, domain-diverse unlabeled speech for cross-lingual self-supervised training. We acquired multimedia content (films, news broadcasts, interviews) from publicly accessible resources for Persian, Arabic, and Urdu, with language verification through platform metadata or manual inspection. Audio tracks were extracted, resampled to 16 kHz, and processed using Silero VAD\footnote{\url{github.com/snakers4/silero-vad.git}} to remove non-speech segments, retaining only chunks with speech probability exceeding 70\%~\cite{Assembly_AI_industrialscalemultilingual}. Audio was segmented into 3-32 second chunks suitable for training, resulting in the pretraining corpus detailed in Table~\ref{tab:train-test}.

\begin{table}
  \centering
    \begin{tabular}{lcc c}
      \hline
      \textbf{Language} & \textbf{Pretraining} & \textbf{Train Set} & \textbf{Test Set} \\
      \hline
      Urdu     & 816 hrs & 60 hrs & 10 hrs \\
      Persian  & 878 hrs & 69 hrs & 11 hrs \\
      Arabic   & 1,310 hrs & 74 hrs & 11 hrs \\
      \hline
    \end{tabular}
  \caption{Data statistics showing pretraining corpus (after VAD-based segmentation) and labeled train/test splits.}
    \vspace{-7.5pt}
  \label{tab:train-test}
\end{table}

\subsection{Labeled Data}

We curated evaluation-ready labeled corpora by combining data from multiple sources to remove bias and ensure diversity across speakers, dialects, and contexts. Sources include Common Voice~\cite{ardila-etal-2020-common}, and other datasets:

\begin{itemize}
    \item \textbf{Urdu}: IndicVoices~\cite{javed2024}, Urdu Speech-To-Text Dataset\footnote{\url{www.kaggle.com/datasets/themohal/tiny-urdu-speech-text-dataset}}
    \item \textbf{Persian}: Persian Speech Corpus\footnote{\url{fa.persianspeechcorpus.com}}, ParsiGoo\footnote{\url{huggingface.co/datasets/Kamtera/ParsiGoo}}, 
    Persian Speech\footnote{\url{github.com/persiandataset/PersianSpeech}},
    ManaTTS-Persian-Speech-Dataset~\cite{qharabagh2024manattspersianrecipecreating}, 
    Persian text-to-speech audio~\cite{nima_moradi_2024} 
    \item \textbf{Arabic}: OpenSLR~\cite{kolobov2021mediaspeechmultilanguageasrbenchmark}, MGB-2~\cite{mgb_2}
\end{itemize}

All audio was resampled to 16 kHz with normalized transcription following~\cite{bandarupalli-etal-2025-towards}. Data was stratified into training and test sets as shown in Table~\ref{tab:train-test}. Detailed breakdowns of the training and validation splits for each dataset are provided in Appendix \ref{sec:data_distribution}.

\begin{figure*}[hbt!]
  \centering
  \includegraphics[width=\linewidth]{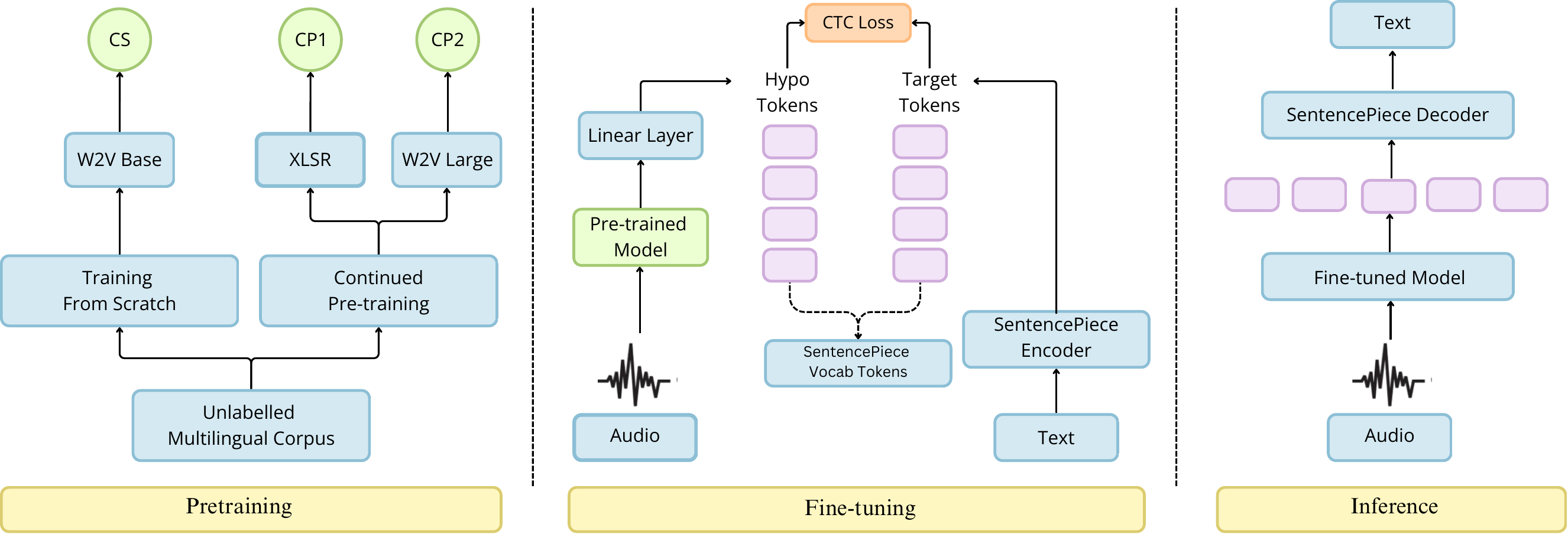}
  \caption{Overview of our experimental framework. The pipeline illustrates three training strategies: CS (Wav2Vec 2.0 Base trained from scratch), CP1 (XLS-R 300M with continuous pretraining), and CP2 (Wav2Vec 2.0 Large with continuous pretraining). All models undergo pretraining on our 3,000-hour multilingual corpus followed by language-specific fine-tuning with SentencePiece tokenization.}
  \vspace{-7.5pt}
  \label{fig:training_pipeline}
\end{figure*}




\section{Experiments}

We design systematic experiments to assess efficient automatic speech recognition (ASR) for low-resource, morphologically complex Persian-Arabic languages. The experimental design is guided by three research questions: \textit{(1) Can cross-lingual unlabeled speech improve ASR performance in low-resource Perso-Arabic languages? (2) Which pretraining initialization strategy yields the best adaptation with continual pretraining? (3) Can compact, parameter-efficient models achieve performance comparable to much larger state-of-the-art systems through strategic pretraining and tokenization?} Figure~\ref{fig:training_pipeline} summarizes our training pipeline.

\subsection{Pretraining Strategies}

To answer these questions, we compare three distinct pretraining initialization strategies:

\textbf{From-Scratch Training (CS):} Wav2Vec~2.0 Base (95M parameters), trained entirely from randomly initialized weights on our 3,000-hour multilingual corpus. This setup isolates the effect of cross-lingual unlabeled data without prior representation learning.

\textbf{Cross-lingual Continual Pretraining (CP1):} XLS-R (300M parameters), pretrained on 436K hours across 128 languages. We further adapt it to our multilingual corpus to examine whether broad multilingual exposure is beneficial for Perso-Arabic languages.

\textbf{English-centric Continual Pretraining (CP2):} Wav2Vec~2.0 Large (300M parameters), pretrained on 65K hours of primarily English speech. This setting tests whether pretraining on a high-resource language can transfer effectively to the target language family.

All three configurations use identical fine-tuning data and hyperparameters to isolate the effect of initialization strategy.

\subsection{Tokenization with SentencePiece}

Perso-Arabic languages exhibit orthographic complexity and rich morphology, making character-level tokenization inadequate. To address this, we integrate \textbf{SentencePiece} with Byte-Pair Encoding (BPE) into the Wav2Vec~2.0 framework. \textbf{Language-specific} subword vocabularies are learned from training transcripts and used to initialize the CTC output layer, enabling direct subword prediction. During inference, subword sequences are decoded into words for final transcription. Further implementation details are provided in Appendix~\ref{sec:sentencepiece}. Empirical comparisons, in Table~\ref{tab:tokenization_comparison}, confirm that SentencePiece-based tokenization yields reductions in WER compared to character-level decoding.

\subsection{Training Procedure}

All models are trained in two stages: (1) \textit{unsupervised pretraining} on the unlabeled corpus, followed by (2) \textit{supervised fine-tuning} on the labeled datasets. Hyperparameters and train/validation splits are standardized across all settings.

\subsubsection{Pretraining}

The CS model is trained from scratch for 200k steps, requiring substantially longer optimization to learn meaningful representations. In contrast, CP1 and CP2 leverage pretrained representations and converge within 40k continual pretraining steps. Across all conditions, early stopping based on validation loss is applied to avoid overfitting during pretraining.

\subsubsection{Fine-tuning}
\begin{table}[h]
  \centering
    \begin{tabular}{lccc}
      \hline
      \textbf{Language} & \textbf{CS} & \textbf{CP1} & \textbf{CP2} \\
      \hline
      Urdu     & 55.8 & 41.1 & 43.2 \\
      Persian  & 28.5 & 21.4 & 19.9 \\
      Arabic   & 71.4 & 49.8 & 58.9 \\
      \hline
    \end{tabular}
  \caption{Validation loss post fine-tuning.}
  \vspace{-12.5pt}
  \label{tab:valid_loss}
\end{table}
Each model is fine-tuned separately for Persian, Arabic, and Urdu for up to 50k steps with early stopping. While WER is our primary evaluation metric, we additionally report \textbf{validation loss} (Table~\ref{tab:valid_loss}) to capture optimization dynamics. Validation loss offers complementary insight into convergence and stability across languages, highlighting that CP1 and CP2 converge substantially faster and to lower minima compared to CS. Final evaluation is performed using WER on held-out test sets (Table~\ref{table:comprehensive_results}).

\subsection{Baseline Fine-tuning}

To contextualize our results, we benchmark against three strong baselines: \textit{Whisper Large v3}, \textit{Seamless Large v2}, and \textit{Wav2Vec~2.0 Large}. These baselines are fine-tuned only on our labeled datasets without additional pretraining on our unlabeled corpus. Whisper and Seamless represent state-of-the-art systems (1.5B+ and 2.3B parameters, respectively), while Wav2Vec~2.0 Large serves as a capacity-matched control.

For fairness, all baselines were fine-tuned until convergence using publicly available scripts, with early stopping triggered once validation loss indicated overfitting. This matches the stopping criteria applied to our models. 
\newline

Full hyperparameter details, along with our training scripts, are released on github for reproducibility. All experiments were conducted on \textbf{NVIDIA A100 80GB GPUs}. 

\begin{table}
  \centering
  \resizebox{\columnwidth}{!}{%
    \begin{tabular}{lccc}
      \hline
      \textbf{Model} & \textbf{Urdu} & \textbf{Persian} & \textbf{Arabic} \\
      \hline
      \multicolumn{4}{c}{\textit{Our Pretrained Models}} \\
      CS (from scratch)       & 25.0 & 25.4 & 38.1 \\
      CP1 (XLS-R init.)          & 22.2 & 22.3 & 35.3 \\
      CP2 (W2V Large init.)      & 20.6 & 17.1 & 32.9 \\
      \hline
      \multicolumn{4}{c}{\textit{Baseline Models}} \\
      
      W2V Large (300 M)      & 39.3 & 33.2 & 48.7 \\
      Seamless Large v2 (2.3B)        & 31.9 & 41.1 & 34.8 \\
      Whisper Large v3 (1.5B)         & 17.2 & 21.4 & 27.2 \\
      \hline
    \end{tabular}
  }
  \caption{Comprehensive WER comparison across all evaluated models after fine-tuning on labeled data.}
  \vspace{-10pt}
  \label{table:comprehensive_results}
\end{table}

\section{Results and Analysis}

We systematically evaluate our approaches to address the three core research questions, presenting quantitative results in Table~\ref{table:comprehensive_results}.

\subsection{Impact of Cross-Lingual Pretraining Data}

The comparison between Wav2Vec 2.0 (W2V) Large and our three models (CS, CP1, CP2) highlights the central role of in-domain unlabeled data. When fine-tuned directly without additional pretraining, W2V Large (300M parameters, 65K hours of English exposure) yields substantially higher WERs across all languages. In contrast, our CS model—trained from scratch on only 3K hours of Perso-Arabic unlabeled audio with 95M parameters—already outperforms W2V Large. This result underscores that domain-relevant pretraining, even at smaller scale, provides greater benefit than model size or large but typologically distant corpora. Both CP1 and CP2 further reduce WERs, confirming the value of targeted continual pretraining.

\subsection{Initialization Strategy Analysis}

Our three initialization strategies exhibit distinct performance trends. As expected, CS, trained from scratch, underperforms relative to CP1 and CP2 due to its lack of prior knowledge and smaller capacity. More interestingly, CP2 consistently outperforms CP1 despite its smaller pretraining corpus. We hypothesize that this difference stems from the distribution of pretraining data: CP2 was initialized from $\sim$65K hours of English-only data, whereas CP1 was initialized from $\sim$65K hours of English combined with $\sim$375K hours of predominantly European languages. Since pretraining fundamentally shapes the acoustic feature extractor, English-only pretraining may yield representations that are more broadly transferable whereas, when the pretraining corpus is heavily dominated by distant languages, the learned representations may become tuned to phonetic patterns common in those distant languages, thereby leading to a negative transfer to target languages. In other words, while pretraining is generally beneficial up to a point, excessively broad exposure to unrelated language families may hinder performance compared to more targeted pretraining.

\subsection{Resource Efficiency}

Our 300M parameter CP2 model demonstrates remarkable efficiency while maintaining competitive performance against much larger state-of-the-art systems. Most notably, our model outperforms Whisper Large v3 on Persian despite using 5× fewer parameters, and achieves performance within 3-6\% WER of Whisper Large v3 for Urdu and Arabic while maintaining significant parameter advantage

The performance gap can be attributed to Whisper's extensive prior exposure to labeled data—739 hours for Arabic and 104 hours for Urdu—compared to our model's training exclusively on our smaller curated dataset. For Persian, where Whisper had minimal prior exposure (24 hours), our targeted pretraining strategy demonstrates clear advantages.

This comparison demonstrates two critical points: (1) effective use of unlabeled data can compensate for limited labeled supervision, and (2) parameter efficiency does not preclude competitive or even superior performance. By exploiting targeted continual pretraining, CP2 provides a low-resource pathway to high-quality ASR, contrasting with the underperformance of much larger systems such as Seamless Large v2 (2.3B parameters), which failed to converge under the same conditions. These findings establish that model scale alone is insufficient—adaptation strategy and data relevance are key to success in low-resource ASR.

\section{Conclusion}

In this work, we present a systematic investigation into resource-efficient ASR for morphologically complex, low-resource languages, with a focus on Perso-Arabic script languages as a representative case study. Our 300M parameter model achieves results competitive with state-of-the-art systems over 5× larger, using substantially less labeled data and computational resources.

Our findings challenge the prevailing assumption that ASR quality scales primarily with model size and data volume. Instead, we show that targeted cross-lingual continuous pretraining, morphologically-aware tokenization, and careful data curation are more critical factors for low-resource languages. Comparing our continual pretraining variants reveals that the relevance of pretraining data, not merely its scale, drives effective transfer learning. This insight suggests that focused, linguistically-informed approaches may be more valuable than broad multilingual exposure for underrepresented languages.

Through our scalable data curation pipeline and strategic utilization of unlabeled multilingual data, we provide a practical pathway toward high-quality ASR that is independent of massive proprietary datasets or computational infrastructure. This work contributes to a more inclusive vision of speech technology where linguistic diversity is not limited by resource constraints, enabling robust ASR systems for the hundreds of millions of speakers of low-resource languages worldwide.

\section*{Limitations}

Our study presents several limitations. First, while we show that cross-lingual pretraining benefits individual low-resource languages, our fine-tuning is performed separately for each language. We have not yet explored the effects of multilingual joint fine-tuning, which could further improve robustness and transfer. Second, although our unlabeled speech corpus is carefully curated for quality and diversity, it may not encompass the full range of dialects, domains, and spontaneous speech present in real-world scenarios. Third, our evaluation is centered on academic benchmark datasets, leaving open questions regarding the practical robustness and end-user acceptance of our models in deployment settings. 

\bibliography{custom}

\appendix

\section{Labeled Data Distribution Across Datasets}
\label{sec:data_distribution}

Most existing studies evaluating Automatic Speech Recognition (ASR) models focus on assessing performance using a single dataset. For instance, models trained on the Common Voice dataset are typically evaluated only on the same dataset. However, when these models are tested on other datasets without fine-tuning, their performance often proves suboptimal, highlighting a lack of generalizability.

To address this limitation and improve the robustness of ASR models across diverse datasets, we adopted a multi-dataset approach for training and evaluation. For each target language, we combined a fraction of data from multiple datasets, including Common Voice, to create comprehensive training and testing splits. This approach ensures that models are exposed to a wider variety of speech patterns, accents, and recording conditions, thereby enhancing their generalizability.

Below, we provide the exact durations of the datasets used for each language, along with the specific train and validation splits, as summarized in Tables~\ref{tab:urdu_splits}, \ref{tab:persian_splits}, and \ref{tab:arabic_splits}.


\begin{table}[h]
  \centering
  \resizebox{\columnwidth}{!}{%
    \begin{tabular}{lcc}
      \hline
      \textbf{Urdu Dataset} & \textbf{Train Duration} & \textbf{Validation Duration} \\
       & \textbf{(hours)} & \textbf{(hours)} \\
      \hline
      Common Voice & 4.23 & 0.73 \\
      SME\_news & 7.48 & 1.28 \\
      Tiny Urdu Speech Corpus & 5.41 & 0.95 \\
      Indic Voices & 42.9 & 7.70 \\
      \hline
    \end{tabular}
  }
  \caption{Urdu dataset splits for fine-tuning.}
  \label{tab:urdu_splits}
\end{table}


\begin{table}[h]
  \centering
  \resizebox{\columnwidth}{!}{%
    \begin{tabular}{lcc}
      \hline
      \textbf{Persian Dataset} & \textbf{Train Duration} & \textbf{Validation Duration} \\
       & \textbf{(hours)} & \textbf{(hours)} \\
      \hline
      Common Voice & 36.5 & 6.44 \\
      Persian Speech Corpus & 2.28 & 0.21 \\
      My Audio Tiny & 2.25 & 0.34 \\
      TTS Female & 22.6 & 4.04 \\
      Moradi & 0.83 & 0.13 \\
      ParsiGOO & 3.56 & 0.64 \\
      \hline
    \end{tabular}
  }
  \caption{Persian dataset splits for fine-tuning.}
  \label{tab:persian_splits}
\end{table}


\begin{table}[h]
  \centering
  \resizebox{\columnwidth}{!}{%
    \begin{tabular}{lcc}
      \hline
      \textbf{Arabic Dataset} & \textbf{Train Duration} & \textbf{Validation Duration} \\
       & \textbf{(hours)} & \textbf{(hours)} \\
      \hline
      Common Voice & 27.5 & 4.89 \\
      SLR-108 & 8.5 & 1.53 \\
      MGB & 37.0 & 5.11 \\
      \hline
    \end{tabular}
  }
  \caption{Arabic dataset splits for fine-tuning.}
  \label{tab:arabic_splits}
\end{table}

\pagebreak
\section{SentencePiece Tokenization}
\label{sec:sentencepiece}

To improve the tokenization process in our speech recognition pipeline, we incorporated SentencePiece tokenization within the Fairseq framework. We first trained a Byte-Pair Encoding (BPE) SentencePiece model with a vocabulary size of 512, using the transcriptions from the training dataset. This vocabulary was subsequently utilized to initialize the Connectionist Temporal Classification (CTC) layer of the wav2vec model.

\subsection{Integration of SentencePiece in Fairseq}

In the default character-based tokenization used in Fairseq, sentences are split into individual characters. To integrate SentencePiece, we modified this process by first segmenting the sentence into words. Instead of directly splitting words at the character level, we applied SentencePiece encoding to tokenize each word into subword units. This approach retains meaningful subword structures while reducing the token sequence length.

During inference, the decoded tokens were grouped at the word level, and each set of tokens was converted back into corresponding words. Finally, the words were concatenated to reconstruct the complete transcription.

\subsection{Impact on Word Error Rate (WER)}

To evaluate the impact of SentencePiece tokenization, we fine-tuned our CP2 model using both the default character-based tokenization and the proposed SentencePiece-based approach. The WER comparison across Urdu, Persian, and Arabic is summarized in Table~\ref{tab:tokenization_comparison}.

\begin{table}[h]
  \centering
  \resizebox{\columnwidth}{!}{%
    \begin{tabular}{lccc}
      \hline
      \textbf{Model} & \textbf{Urdu} & \textbf{Persian} & \textbf{Arabic} \\
      \hline
      CP2 (Character-based) & 25.8 & 26.2 & 39.0 \\
      CP2 (SentencePiece-based) & 20.6 & 17.1 & 32.9 \\
      \hline
    \end{tabular}
  }
  \caption{Word Error Rate (WER) comparison between character-based and SentencePiece-based tokenization.}
  \label{tab:tokenization_comparison}
\end{table}

The results demonstrate a significant reduction in WER across all three languages, highlighting the effectiveness of subword-level tokenization over character-based tokenization in CTC-based speech recognition.
\end{document}